\documentclass[runningheads]{llncs}

\usepackage[T1]{fontenc}
\usepackage{lmodern}
\usepackage{graphicx}
\usepackage{amsmath}
\usepackage{amssymb}
\usepackage{booktabs}
\usepackage{array}
\usepackage[caption=false]{subfig}
\usepackage{url}

\begin{document}

\title{From Mono to Stereo: Accelerating Binocular Gaussian Splatting via Reprojection and Selective Patching}
\titlerunning{From Mono to Stereo}

\author{Hongfei Zhu\inst{1,2} \and
Ling Zhou\inst{2}\thanks{Corresponding author.}}
\authorrunning{H. Zhu and L. Zhou}
\institute{Shanghai Jiao Tong University, Shanghai, China\\
\email{vit\_oo@sjtu.edu.cn} \and
Alpha Labs, Goertek, Shanghai, China\\
\email{ling.zhou@goertek.com}}

\maketitle

\begin{abstract}
Binocular rendering requires two nearby views of the same scene and therefore repeats substantial visibility and shading work.
We present a 2D Gaussian Splatting (2DGS) pipeline that fully renders a dominant-eye RGB image and an alpha-weighted depth proxy, reprojects that image to the affiliated eye, and repairs uncovered pixels.
Small interior gaps are interpolated, whereas larger disoccluded regions are identified as regions of interest (ROIs) and selectively re-rendered.
The depth proxy reuses the alpha-blending weights computed during dominant-eye rasterization, avoiding a separate depth-rendering pass.
An adaptive ROI generator localizes the required updates using reprojected image boundaries and optional connected center-hole detection.
On DTU, Tanks and Temples, and MipNeRF-360, the method reduces the measured time of a sequential two-pass binocular reference by 15.5\% to 28.8\% and peak GPU memory by 6\% to 11\%.
The corresponding affiliated-eye quality degradation is at most 1.3~dB PSNR, 0.02 SSIM, and 0.02 LPIPS, representing a measurable trade-off that requires application-specific perceptual validation.
These results establish a practical efficiency--quality trade-off for controlled static-scene stereo rendering and motivate future evaluation under continuous motion and on physical VR hardware.

\keywords{Virtual Reality \and Gaussian Splatting \and Binocular Rendering \and Rasterization Acceleration}
\end{abstract}

\section{Introduction}

Virtual reality (VR) systems render two horizontally displaced views so that the visual system can recover binocular depth cues~\cite{howard2012perceiving}.
The two views overlap substantially, but straightforward stereo pipelines rasterize each eye independently and repeat much of the same work~\cite{hu2025efficient,wissmann2020accelerated}.
This repeated cost is important for high-resolution head-mounted displays, mobile rendering, and interactive telepresence, where both latency and memory budgets are constrained.

Image reprojection can reuse samples across time or viewpoints, but disocclusions expose content that is absent from the source image.
Temporal antialiasing systems address a related problem through history validation and reconstruction~\cite{yang2020survey}; stereo methods instead transfer samples between the two eye views within the same frame~\cite{yang2025accelerating,wissmann2020accelerated,chen2025yoro}.
For Gaussian Splatting, the depth required by this transfer can be accumulated during rasterization, while missing regions can be re-rendered directly from the scene representation.

We therefore study a hybrid binocular pipeline for static 2DGS scenes.
The dominant eye is fully rendered with RGB and an alpha-weighted depth proxy, and its image is reprojected to the affiliated eye using the known relative pose.
The resulting holes are divided into small gaps and larger disoccluded ROIs; interpolation handles the former, and selective 2DGS rasterization handles the latter.
Our contributions are:

\begin{itemize}
    \item A reprojection-and-patch pipeline that avoids a complete second-eye rasterization by combining depth-guided warping, hole categorization, interpolation, and selective ROI rendering.
    \item A joint RGB-depth rendering component that reuses the alpha-blending weights already computed by the 2DGS rasterizer, avoiding a separate depth-rendering pass.
    \item Evaluation on three real-world multi-view benchmarks under a controlled static-scene protocol, with per-scene timing and memory analysis.
\end{itemize}

\begin{figure}[!htbp]
    \centering
    \includegraphics[width=0.92\textwidth]{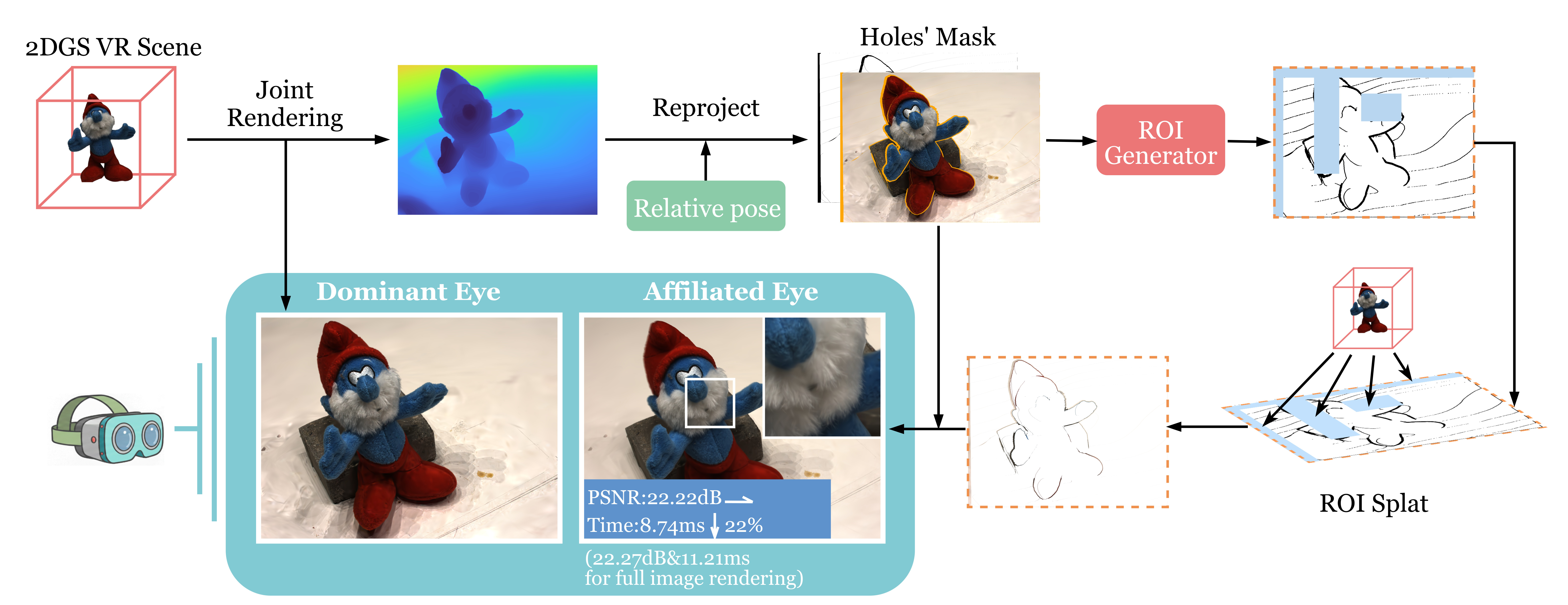}
    \caption{Overview of the binocular rendering pipeline. The dominant eye is fully rasterized to obtain an RGB image and depth proxy, which are reprojected to the affiliated eye. Small gaps are interpolated, while larger regions of interest (ROIs) are selectively re-rendered.}
    \label{fig:pipeline}
\end{figure}

\section{Related Work}

\subsection{Temporal and Stereo Reprojection}

Reprojection reuses previously shaded samples by transforming them with depth, camera motion, or both.
Asynchronous reprojection warps an earlier frame to compensate for updated head pose~\cite{google2018asynchronous}.
Temporal antialiasing (TAA) and temporal upsampling extend this principle by accumulating samples over time and rejecting invalid history; the survey by Yang et al.~\cite{yang2020survey} summarizes reconstruction, history validation, and the characteristic failure modes around disocclusions and motion.
These temporal techniques motivate our treatment of invalid samples, but our main transfer is spatial: it reuses the dominant-eye image for the affiliated eye within the same binocular frame.

Stereo reprojection has previously been combined with temporal supersampling~\cite{yang2025accelerating} and with adaptive ray tracing~\cite{wissmann2020accelerated}.
Wissmann et al. forward-warp one eye and trace rays only where reprojection fails, which is conceptually close to our selective repair strategy; our pipeline instead operates on an explicit 2DGS scene and repairs selected regions through Gaussian rasterization.
YORO~\cite{chen2025yoro} synthesizes a second mobile-VR view using reprojection and filter-based completion without re-rendering newly visible content.
Our qualitative comparison in Fig.~\ref{fig:yoro} illustrates why this reprojection-only design can distort near-scene disocclusions.
Because YORO uses a Unity rasterization pipeline rather than a 2DGS renderer, we treat it as a qualitative reference rather than report implementation-dependent timing comparisons.

\subsection{Gaussian Splatting Efficiency and View Consistency}

3D Gaussian Splatting (3DGS) introduced a visibility-aware tile rasterizer for real-time novel-view synthesis~\cite{kerbl2023gaussian}.
Subsequent work reduces representation and rendering cost by pruning Gaussians and compressing their attributes.
Compact 3DGS uses learnable masks and quantized attributes~\cite{lee2024compact}, while EAGLES combines quantized embeddings with coarse-to-fine optimization and pruning~\cite{girish2024eagles}.
These methods reduce the cost of rendering each view and are complementary to our reuse of samples across a stereo pair.

View consistency is also relevant to reprojection.
StopThePop improves per-ray ordering and evaluates artifacts that become visible during viewpoint motion~\cite{radl2024stopthepop}.
Our method does not modify the underlying 2DGS ordering; instead, it relies on the depth proxy produced by the rasterizer and explicitly re-renders disoccluded ROIs.
This distinction also exposes a limitation: view-dependent appearance and temporal popping inherited from the source view are not corrected by image warping alone.

\subsection{Dynamic Gaussian Scenes}

Dynamic Gaussian representations model time-varying geometry and appearance, for example through learned deformations in 4D Gaussian Splatting~\cite{wu2024fourdimensional}.
The present work assumes a fixed 2DGS scene and evaluates isolated stereo pairs.
Supporting moving objects would require a time-dependent Gaussian representation, current per-frame depth, and temporal validation of reprojected samples.
Continuous head motion in an otherwise static scene also requires temporal evaluation because ROI boundaries and interpolation decisions may change between frames.

\section{Method}

\subsection{Pipeline Overview}

As illustrated in Fig.~\ref{fig:pipeline}, our method replaces the conventional independent binocular rendering pipeline with a more efficient hybrid approach that exploits the geometric correlation between the two eye views.
The overall pipeline consists of four main stages.

\paragraph{Stage 1: Joint RGB-Depth Rendering of the Dominant Eye.}
The pipeline begins by performing a full rasterization of the dominant eye, e.g., the left eye, from the 2D Gaussian Splatting scene.
During front-to-back compositing, splat $i$ contributes with the standard alpha-blending weight $w_i=\alpha_iT_i$.
We reuse this weight in the same CUDA kernel to accumulate the depth proxy $D=\sum_i d_iw_i$ alongside RGB, avoiding a separate depth pass.
This alpha-weighted depth supports the subsequent reprojection, although it may differ from physical surface depth in multi-layer, semi-transparent, or strongly view-dependent regions.

\paragraph{Stage 2: Depth-Guided Reprojection.}
Using the rendered depth map and the known relative pose between the two eyes, we reproject the dominant-eye image to synthesize an initial view for the affiliated eye, e.g., the right eye.
Each pixel in the dominant eye is unprojected into 3D world coordinates using its depth value and the source camera parameters, and then projected onto the affiliated-eye image plane.
This step produces a largely complete image for the second eye at minimal computational cost, since it involves only per-pixel geometric transformations rather than full Gaussian rasterization.

\paragraph{Stage 3: Hole Categorization.}
The reprojected image inevitably contains uncovered pixels, referred to as holes, arising from disocclusions and field-of-view differences between the two viewpoints.
We classify these holes into two types: gaps, which are small, scattered interior pixels caused by depth discontinuities and parallax-induced disocclusions; and ROIs, which comprise peripheral areas visible only in the affiliated eye, large connected gap clusters caused by substantial parallax, and a conservative buffer around these regions.

\paragraph{Stage 4: Hybrid Hole Patching.}
The two types of holes are patched using different strategies tailored to their characteristics.
Gaps, which are typically small and surrounded by valid reprojected pixels, are efficiently filled through adaptive interpolation.
ROIs, including peripheral areas with genuinely new scene content and large connected center-hole clusters, are resolved by an adaptive ROI generator that identifies bounded regions for targeted rasterization.

\subsection{Why 2D Gaussian Splatting}

Our pipeline supports any Gaussian Splatting variant that outputs per-pixel depth, but accurate geometry is crucial because depth errors cause warping artifacts and additional holes.
Table~\ref{tab:geometry} compares three representative backbones.
The volumetric ellipsoids of 3DGS~\cite{kerbl2023gaussian} provide relatively coarse surfaces, while SuGaR~\cite{guedon2024sugar} improves geometry through surface alignment.
2D Gaussian Splatting (2DGS)~\cite{huang2024twodgs} instead uses oriented planar disks on tangent planes, yielding the best DTU and Tanks and Temples geometry scores among the compared methods.
We therefore adopt 2DGS for its accurate, view-consistent geometry.

\begin{table}[t]
    \centering
    \caption{Quantitative comparison of geometry reconstruction quality.}
    \label{tab:geometry}
    \begin{tabular}{lcc}
        \toprule
        Method & DTU (CD) & T\&T (F1) \\
        \midrule
        3DGS~\cite{kerbl2023gaussian} & 1.96 & 0.09 \\
        SuGaR~\cite{guedon2024sugar} & 1.33 & 0.19 \\
        2DGS~\cite{huang2024twodgs} & 0.80 & 0.32 \\
        \bottomrule
    \end{tabular}
\end{table}

\subsection{2D Gaussian Modeling}

A 2D Gaussian primitive has center $\mathbf{p}_k \in \mathbb{R}^3$, tangent vectors $\{\mathbf{t}_u,\mathbf{t}_v\}$, and in-plane scales $\mathbf{s}=(s_u,s_v)$.
Its rotation quaternion gives $\mathbf{R}=[\mathbf{t}_u,\mathbf{t}_v,\mathbf{t}_w]$, where $\mathbf{t}_w=\mathbf{t}_u \times \mathbf{t}_v$ is the normal.

The geometry of the 2D Gaussian in world space is described by a homogeneous transformation matrix:
\begin{equation}
\mathbf{H} =
\begin{bmatrix}
s_u\mathbf{t}_u & s_v\mathbf{t}_v & \mathbf{0} & \mathbf{p}_k \\
0 & 0 & 0 & 1
\end{bmatrix}.
\end{equation}
For local coordinate $\mathbf{u}=(u,v)$, $\mathbf{P}(u,v)=\mathbf{H}[u,v,0,1]^T$ and
\begin{equation}
G(\mathbf{u})=\exp\left(-\frac{u^2+v^2}{2}\right).
\end{equation}

For perspective-accurate splatting, 2DGS transforms each pixel ray's two homogeneous planes into the primitive frame and computes their intersection on the disk.

\subsection{Reprojection}

Reprojection establishes dense correspondence between stereo views by lifting image pixels into 3D and reprojecting them into another view under known camera geometry.
We adopt a unified pinhole camera model with projection and unprojection operators.

Let $\mathbf{X}=[X,Y,Z]^T \in \mathbb{R}^3$ denote a 3D point expressed in the camera coordinate system.
The perspective projection operator $\pi(\cdot)$ is defined as
\begin{equation}
\pi(\mathbf{X}) =
\mathbf{K}
\begin{bmatrix}
X/Z\\
Y/Z\\
1
\end{bmatrix},
\end{equation}
where $\mathbf{K}\in\mathbb{R}^{3\times 3}$ is the camera intrinsic matrix, and $Z$ denotes the depth of the point along the camera optical axis.
Given a source pixel $\mathbf{u}_{src}$ with depth $d$, its 3D position in the world coordinate system is obtained via unprojection:
\begin{equation}
\mathbf{P}_{world}=\mathbf{R}_{src}\pi^{-1}(\mathbf{u}_{src},d)+\mathbf{t}_{src},
\end{equation}
where $\mathbf{u}_{src}\in\mathbb{R}^3$ is a homogeneous image coordinate, $\pi^{-1}(\cdot)$ denotes the unprojection operator, and $\mathbf{R}_{src}\in SO(3)$ and $\mathbf{t}_{src}\in\mathbb{R}^3$ represent the rotation matrix and translation vector of the source camera pose in the world coordinate system.
The point is then transformed into the destination camera frame and projected onto the target image plane:
\begin{equation}
\mathbf{u}_{dst}=\pi_{dst}\left(\mathbf{R}_{dst}^{T}(\mathbf{P}_{world}-\mathbf{t}_{dst})\right),
\end{equation}
where $\mathbf{R}_{dst}\in SO(3)$ and $\mathbf{t}_{dst}\in\mathbb{R}^3$ denote the rotation and translation of the destination camera, and $\pi_{dst}(\cdot)$ uses the destination camera intrinsics.

\subsection{Adaptive ROI Generator}

\begin{figure}[!htbp]
    \centering
    \begin{minipage}[c]{0.49\textwidth}
        \centering
        \subfloat[Configured ROI paths for far\\ and near objects.\label{fig:roi_paths}]{%
            \includegraphics[width=\linewidth]{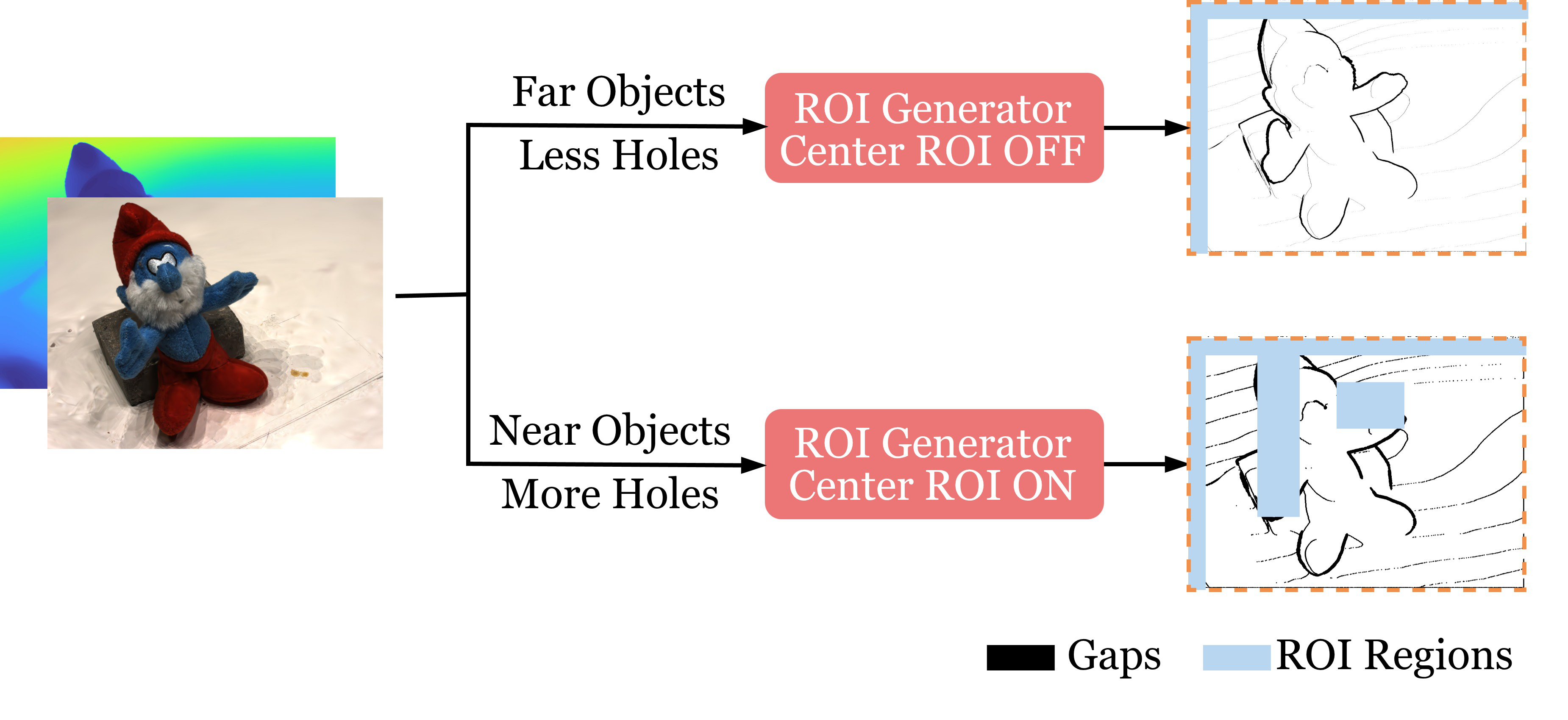}}
    \end{minipage}\hfill
    \begin{minipage}[c]{0.49\textwidth}
        \centering
        \subfloat[Adaptive edge-ROI\\ generation.\label{fig:roi_generation}]{%
            \includegraphics[width=\linewidth]{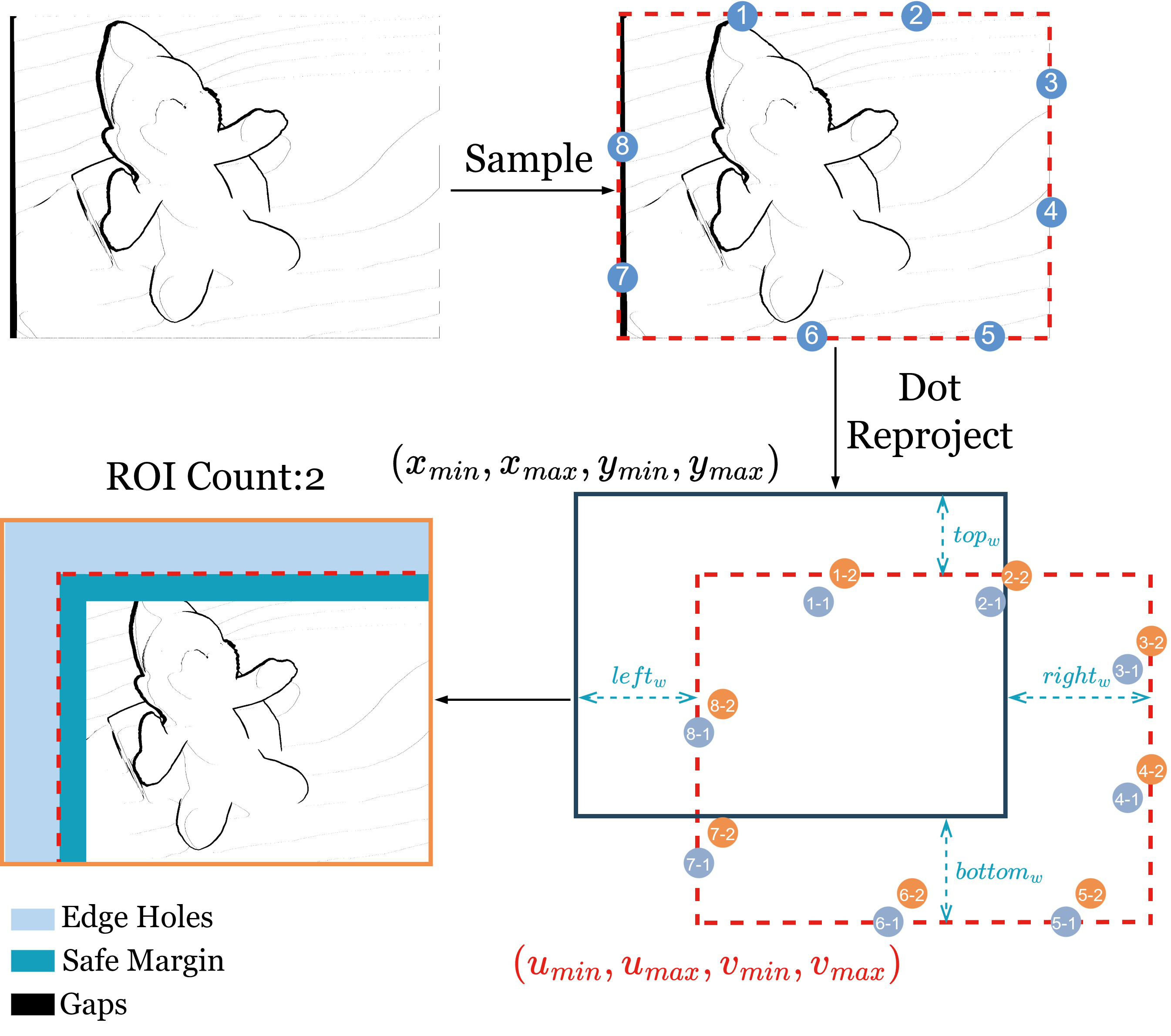}}
    \end{minipage}
    \caption{ROI generation. (a) Step 1 generates edge ROIs and fills remaining small holes by interpolation; for scenes expected to contain close foreground geometry, the user can enable Step 2 to repair connected center-hole regions. (b) Boundary reprojection and tile alignment determine the border regions that require re-rendering.}
    \label{fig:roi_overview}
\end{figure}

Reprojection produces two types of holes.
Gaps are small, scattered interior holes caused by depth variation and parallax, and are filled by interpolation.
ROIs comprise peripheral regions visible only to the affiliated eye and large connected disocclusions near foreground boundaries; together with a small safety margin, they are repaired by targeted rendering.
To limit affiliated-eye rasterization, our adaptive ROI generator restricts this rendering to the required regions, as summarized in Fig.~\ref{fig:roi_overview}.
As illustrated in Fig.~\ref{fig:roi_paths}, far objects generally produce fewer and smaller holes, whereas near objects produce larger interior disocclusions because of greater parallax.
The edge-ROI step is therefore always applied, while the optional center-ROI step is enabled by the user before rendering for scenes expected to contain close foreground geometry and wide interior disocclusions.

\paragraph{Step 1: Edge ROI Generation.}
As illustrated in Fig.~\ref{fig:roi_generation}, our ROI generator dynamically identifies the border regions corresponding to edge holes that require re-rendering.
Let $\Omega_{bound}$ denote the boundary of the source image domain.
We uniformly sample $N$ pixels along each image edge.
For each sampled boundary pixel $\mathbf{u}_{src}$, we reproject it at two depth hypotheses $\{d_{min},d_{max}\}$ using Eqs.~(3)--(5).
Let $\mathcal{P}$ denote the set of all such valid projected points in the affiliated eye, where each point $\mathbf{p}\in\mathcal{P}$ implies a coordinate pair $(x,y)$.

We then compute the axis-aligned bounding box by finding the coordinate extrema over the set $\mathcal{P}$:
\begin{equation}
\begin{aligned}
x_{min} &= \min_{(x,y)\in\mathcal{P}} x, & x_{max} &= \max_{(x,y)\in\mathcal{P}} x,\\
y_{min} &= \min_{(x,y)\in\mathcal{P}} y, & y_{max} &= \max_{(x,y)\in\mathcal{P}} y.
\end{aligned}
\end{equation}
Based on the projected boundary samples, we estimate the uncovered border regions in the affiliated eye.
Let $(u_{min},u_{max},v_{min},v_{max})$ denote the resulting valid projection bounds.
Given the image width $W$ and height $H$, we compute the border widths along four sides by expanding the valid projection range with a fixed edge margin $m$.
Specifically, the left and right border widths are defined as:
\begin{equation}
left_w=
\begin{cases}
\mathrm{clamp}(\lceil u_{min}\rceil+m,0,W/2), & u_{min}>0,\\
0, & \mathrm{otherwise},
\end{cases}
\end{equation}
\begin{equation}
right_w=
\begin{cases}
\mathrm{clamp}((W-1)-\lfloor u_{max}\rfloor+m,0,W/2), & u_{max}<W-1,\\
0, & \mathrm{otherwise}.
\end{cases}
\end{equation}
The top and bottom border heights are computed analogously from $v_{min}$ and $v_{max}$.
Here, $\mathrm{clamp}(x,a,b)=\min(\max(x,a),b)$ limits the ROI size to at most half of the image extent.

After obtaining the coarse projection bounds through boundary reprojection, we align the bounds to the rasterizer's native $16\times16$-pixel tile boundaries, whose size is determined by the original gaussian rasterizer and is therefore not tuned as a method-specific parameter.
Here a margin of $m=32$ pixels was selected empirically during system development as a conservative buffer against small inaccuracies introduced by projection, discretization, and temporal variation, and was subsequently kept fixed across all evaluated scenes and datasets.
The margin controls a coverage--efficiency trade-off: reducing it may produce smaller ROIs and lower rendering cost, but increases the risk of excluding relevant boundary pixels; increasing it improves tolerance to projection errors but causes additional pixels to be rendered and therefore reduces the performance benefit.

\paragraph{Step 2: Center ROI Generation.}
When the user enables center-ROI rendering, we perform an additional center ROI generation step after edge ROI rendering.
We first identify center hole pixels in the reprojected image, i.e., interior pixels that remain unfilled after reprojection and interpolation.
Connected component analysis is then applied to group adjacent center hole pixels into clusters.
For each resulting cluster, we compute an axis-aligned bounding box encompassing the cluster.
The bounding box is expanded by the same margin $m$ and aligned to the tile grid, producing a center ROI.
These center ROIs are rendered via the same targeted ROI rendering pipeline used for edge ROIs, ensuring high-quality hole repair through 2DGS re-rasterization.
The choice to enable Step 2 is made before the rendering run; it is not triggered by an area threshold or by an automatic near/far classifier.
This user-configured choice and the fixed margin are heuristic design settings rather than learned or optimized parameters, and their sensitivity is not evaluated in the current experiments.

\section{Experiment}

The whole pipeline is run on a desktop equipped with an NVIDIA RTX 4090D GPU.

\subsection{Datasets}

We evaluate our method on three widely used multi-view datasets: DTU~\cite{jensen2014large}, a controlled indoor benchmark from which we select 15 scans (24, 37, 40, 55, 63, 65, 69, 83, 97, 105, 106, 110, 114, 118, and 122); Tanks and Temples~\cite{knapitsch2017tanks}, containing eight large-scale indoor and outdoor scenes; and MipNeRF-360~\cite{barron2022mipnerf360}, containing eight high-resolution unbounded scenes with complex geometry and wide baselines. We use an approximate 2:1 training--test split for every scene, yielding 33 training and 16 test images per DTU scan, while the exact counts for the other datasets vary with scene size.

\subsection{Stereo Pair Construction and Evaluation Targets}

Each test sample consists of a dominant eye and an affiliated eye.
The pose of the affiliated eye is first generated by COLMAP.
The pose of the dominant eye is generated with a small viewpoint difference from the affiliated eye while preserving the same orientation, thereby simulating binocular parallax.
The affiliated eye image is compared against the ground truth for evaluating PSNR, SSIM, and LPIPS.
Although no captured reference image exists for the dominant eye, none is needed for the comparison: both rendering strategies use the same full 2DGS rendering at the dominant eye and differ only in how they generate the affiliated eye image.
Consequently, all image-quality metrics are computed at the affiliated eye against the captured image.

\subsection{Experimental Pipeline}

For each scene, COLMAP~\cite{schoenberger2016sfm} recovers camera intrinsics and extrinsics, and a 2DGS model is trained for 30,000 iterations on the training images.
After training, the model is fixed for all rendering and evaluation.

With the stereo pairs prepared, we compare two sequential rendering strategies for producing the binocular frames.
Both strategies begin with exactly the same complete dominant-eye rendering, which produces the dominant-eye RGB image and depth proxy.
The full-rendering reference then performs a second complete rasterization at the affiliated eye pose, whereas our method replaces this second full-image pass with reprojection, interpolation, and selective ROI rasterization.
The reported time for each strategy is the accumulated cost of these sequential stages for the complete stereo pair; neither strategy uses batched multiview rendering or concurrent CUDA streams.
Comparisons with batched or concurrent execution are outside the scope of the reported measurements.
In our method, the shared dominant-eye depth proxy is used, together with the known relative pose, to reproject the dominant-eye image into the viewpoint of the affiliated eye pose.
The reprojected result contains holes from disocclusions and field-of-view differences; gaps are filled by interpolation, while ROI regions are resolved through targeted ROI rendering.

Finally, visual quality is evaluated only on the affiliated eye pose.
Specifically, both the full-rendering baseline at the affiliated eye pose and our generated affiliated eye image are compared against the captured affiliated eye ground-truth image.
Rendering time and GPU memory are measured for the complete stereo pair: the identical dominant-eye pass followed by a complete affiliated eye pass for the reference, and the identical dominant-eye pass followed by reprojection and patching for our method.
All speedup statements below are therefore relative to this sequential two-pass reference.

\subsection{Visual Comparison}

\begin{figure}[!t]
    \centering
    \includegraphics[width=0.72\textwidth]{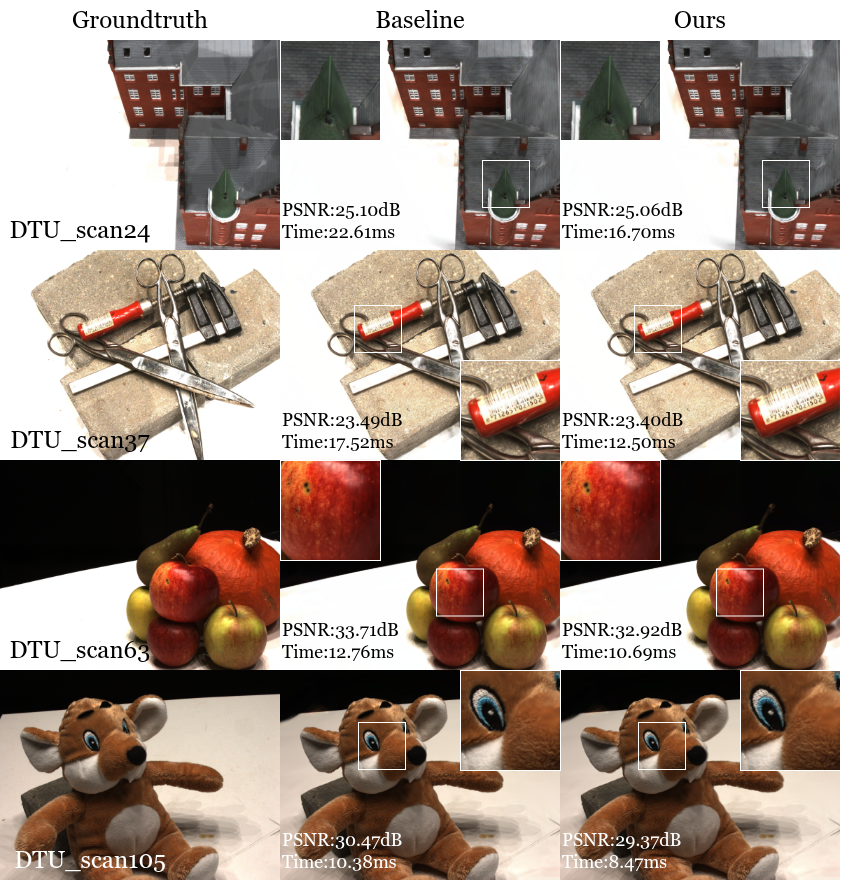}
    \caption{Visual comparison on the DTU dataset.}
    \label{fig:vis_dtu}
\end{figure}

Figure~\ref{fig:vis_dtu} compares the affiliated eye views on DTU: captured ground truth, full-rendered baseline, and the result of our method.
Additional comparisons on Tanks and Temples and MipNeRF-360 are provided in Figs.~\ref{fig:vis_tnt} and~\ref{fig:vis_mipnerf}, respectively, in Appendix~\ref{sec:supp_vis}.
Zoomed-in insets highlight local details.
The examples show that most scene content is preserved, while local differences remain near disocclusions and fine structures.

\subsection{Quantitative Evaluation}

Table~\ref{tab:dataset_average} shows a consistent quality--efficiency trade-off relative to sequential two-pass rendering: affiliated-eye PSNR decreases by 0.82--1.20~dB, while stereo-pair rendering time and GPU memory decrease by 15.5--28.8\% and 6--11\%, respectively.
We report the quality change without interpreting it as perceptually negligible because no headset-based perceptual threshold was evaluated.

\begin{table}[t]
    \centering
    \caption{Per-dataset average quality and efficiency metrics. Quality metrics are computed on the affiliated eye pose; time and GPU memory are measured for the complete stereo pair.}
    \label{tab:dataset_average}
    \resizebox{\textwidth}{!}{
    \begin{tabular}{llccccc}
        \toprule
        Dataset & Method & PSNR (dB) & SSIM & LPIPS & Time (ms) & GPU (MB) \\
        \midrule
        DTU~\cite{jensen2014large} & Baseline & 28.83 & 0.8738 & 0.3730 & 17.81 & 744 \\
        DTU~\cite{jensen2014large} & Ours & 28.01 & 0.8677 & 0.3790 & 15.05 & 661 \\
        \midrule
        Tanks \& Temples~\cite{knapitsch2017tanks} & Baseline & 20.99 & 0.7383 & 0.4145 & 35.67 & 1792 \\
        Tanks \& Temples~\cite{knapitsch2017tanks} & Ours & 20.00 & 0.7219 & 0.4266 & 25.40 & 1681 \\
        \midrule
        MipNeRF-360~\cite{barron2022mipnerf360} & Baseline & 26.14 & 0.7895 & 0.3437 & 43.10 & 2463 \\
        MipNeRF-360~\cite{barron2022mipnerf360} & Ours & 24.94 & 0.7717 & 0.3627 & 30.96 & 2310 \\
        \bottomrule
    \end{tabular}}
\end{table}

\begin{figure}[!t]
    \centering
    \begin{minipage}[c]{0.49\textwidth}
        \centering
        \subfloat[Per-scene scaling trends.\label{fig:corr}]{%
            \includegraphics[width=\linewidth]{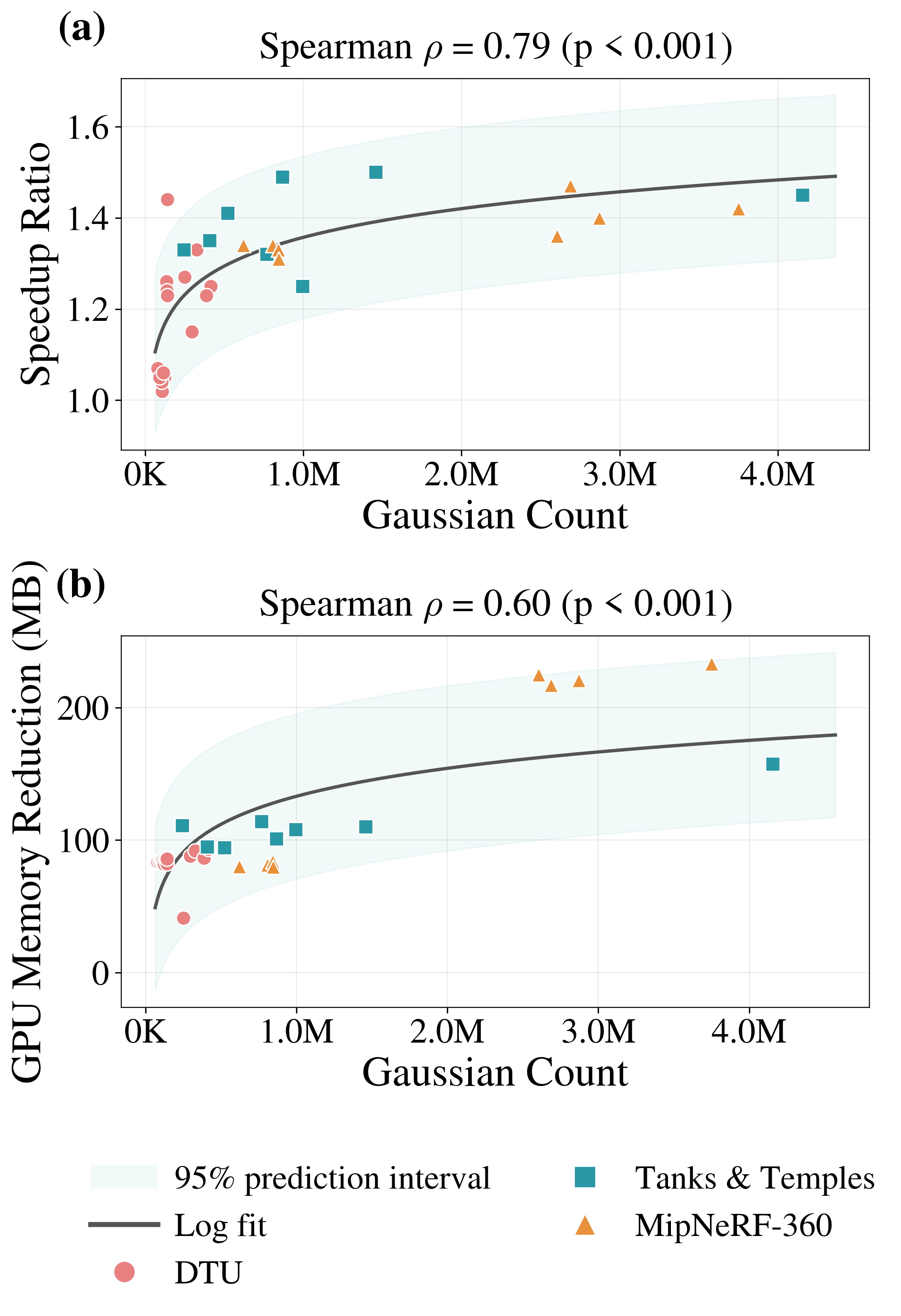}}
    \end{minipage}\hfill
    \begin{minipage}[c]{0.49\textwidth}
        \centering
        \subfloat[Per-dataset time breakdown.\label{fig:timebreakdown}]{%
            \includegraphics[width=\linewidth]{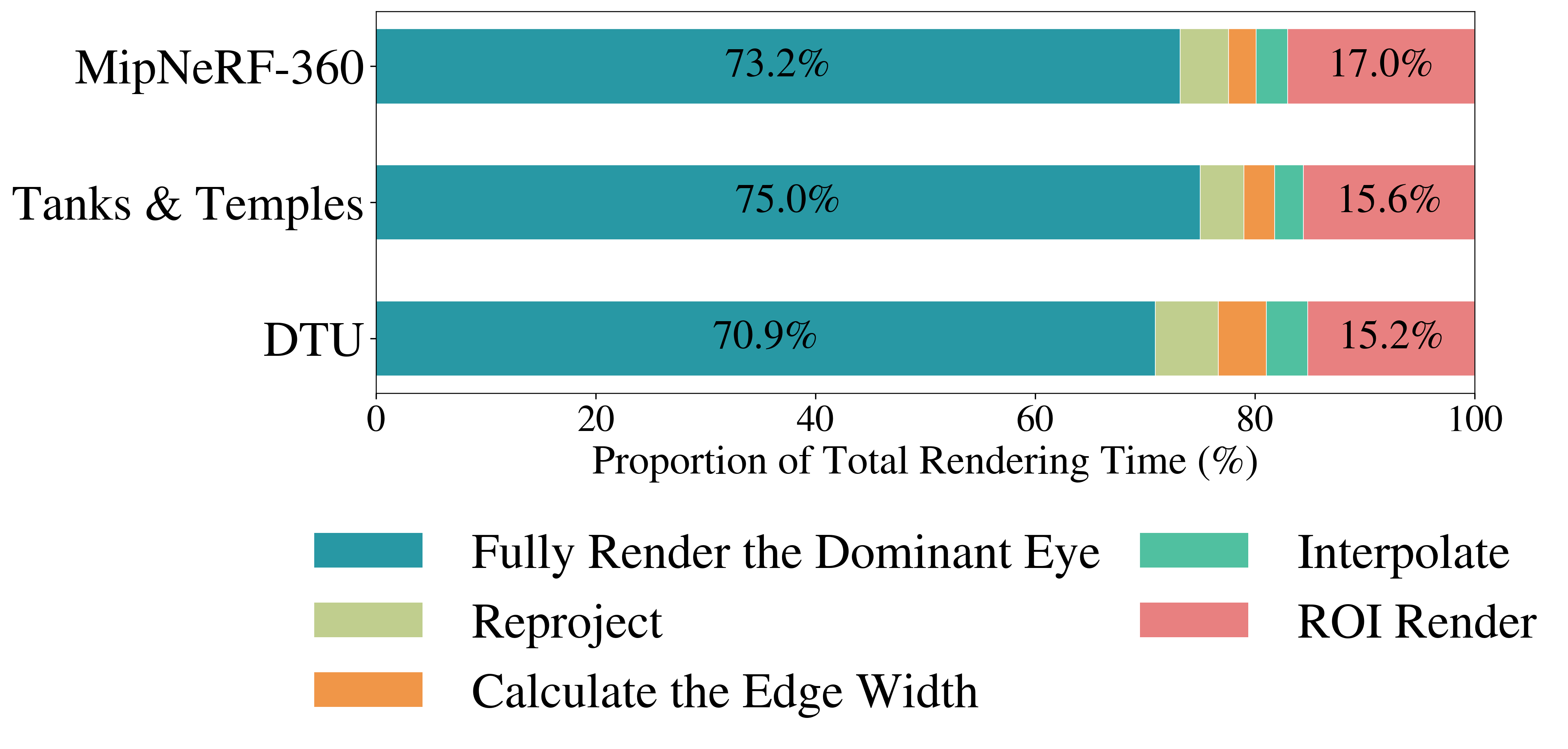}}
    \end{minipage}
    \caption{Efficiency analysis across the evaluated datasets. (a) Per-scene analysis on DTU, Tanks and Temples, and MipNeRF-360; both speedup and GPU memory reduction increase with scene complexity. (b) Per-dataset time breakdown of the proposed pipeline.}
    \label{fig:efficiency_analysis}
\end{figure}

Figure~\ref{fig:efficiency_analysis} summarizes the efficiency analysis across datasets. Specifically, Fig.~\ref{fig:corr} shows that denser scenes generally yield greater speedup and absolute memory savings, with Spearman $\rho=0.79$ and $0.60$, respectively (both $p<0.001$).
These correlations are descriptive rather than causal because Gaussian count is not isolated from resolution, dataset, geometry, or ROI coverage.

Figure~\ref{fig:timebreakdown} identifies dominant-eye rendering as the main bottleneck (about 71--75\% of total time), followed by ROI rendering (about 15--17\%); the remaining stages together account for less than 14\%.

\subsection{Discussion}

As reported in Table~\ref{tab:dataset_average}, the method trades affiliated-eye image quality for lower rendering time and memory use relative to the sequential two-pass reference.
Figure~\ref{fig:corr} further shows that both speedup and GPU memory savings grow with the per-scene Gaussian count, with Spearman $\rho=0.79$ and 0.60, respectively, and both $p<0.001$.
This association is consistent with the cost of full rasterization increasing with primitive count while much of the reprojection overhead is image based.
However, Gaussian count, image resolution, scene geometry, and ROI coverage vary together in the current benchmarks, so the analysis does not establish a causal scaling law.
The time breakdown in Fig.~\ref{fig:timebreakdown} also confirms that the total cost is dominated by rendering the dominant eye.

Several limitations remain.
Our pipeline assumes accurate per-pixel depth; errors directly cause reprojection holes and misaligned pixels, especially in sparse or large-scale outdoor scenes.
Because the method reuses dominant-eye pixels, view-dependent appearance such as specular highlights or reflections may also be inherited from the dominant-eye viewpoint.
In addition, our evaluation uses controlled stereo pairs derived from COLMAP poses on an RTX 4090D desktop.
We have not evaluated perceptual comfort, binocular rivalry, or latency on a physical VR headset.
The timing results compare two sequential pipelines with an identical first-eye pass and do not establish an advantage over batched multiview or concurrent stereo rendering.

The method targets static scenes.
Moving objects would require a time-dependent scene representation, a current depth estimate, and validity tests that prevent stale dominant-eye samples from being transferred to the affiliated eye.
Even with static geometry, continuous camera motion may cause ROI boundaries and interpolation decisions to change between frames; temporal stability therefore requires sequence-level evaluation and may require additional history constraints.
Our current experiments evaluate independent stereo pairs and do not measure this behavior.

\section{Conclusion}

We presented a binocular rendering acceleration framework for static 2D Gaussian Splatting scenes.
The method replaces redundant full rendering of the second eye with depth-guided reprojection and targeted hole filling, using interpolation for gaps and ROI rendering for larger missing regions.

Experiments on DTU, Tanks and Temples, and MipNeRF-360 show reductions of up to 28.8\% in rendering time and 6\% to 11\% in peak GPU memory relative to a sequential two-pass reference, with affiliated-eye changes within 1.3~dB PSNR, 0.02 SSIM, and 0.02 LPIPS.
These results establish the trade-off for controlled static-scene pairs, but do not establish physically calibrated headset performance or temporal perceptual quality.
Future work will evaluate normalized stereo baselines, multiview and concurrent rendering references, parameter sensitivity, continuous head motion, physical VR hardware, and dynamic Gaussian scenes.

\begin{credits}
\subsubsection{\ackname}
The authors thank Goertek for providing the hardware resources and experimental facilities that supported this work.
OpenAI Codex and Anthropic Claude were used solely for grammar editing and language polishing. All research content, technical contributions, and scientific writing are entirely the authors' own work.

\subsubsection{\discintname}
The authors have no competing interests to declare that are relevant to the content of this article.
\end{credits}

\appendix
\section{Supplementary Visual Comparison}
\label{sec:supp_vis}

\begin{figure}[!htbp]
    \centering
    \includegraphics[width=0.72\textwidth]{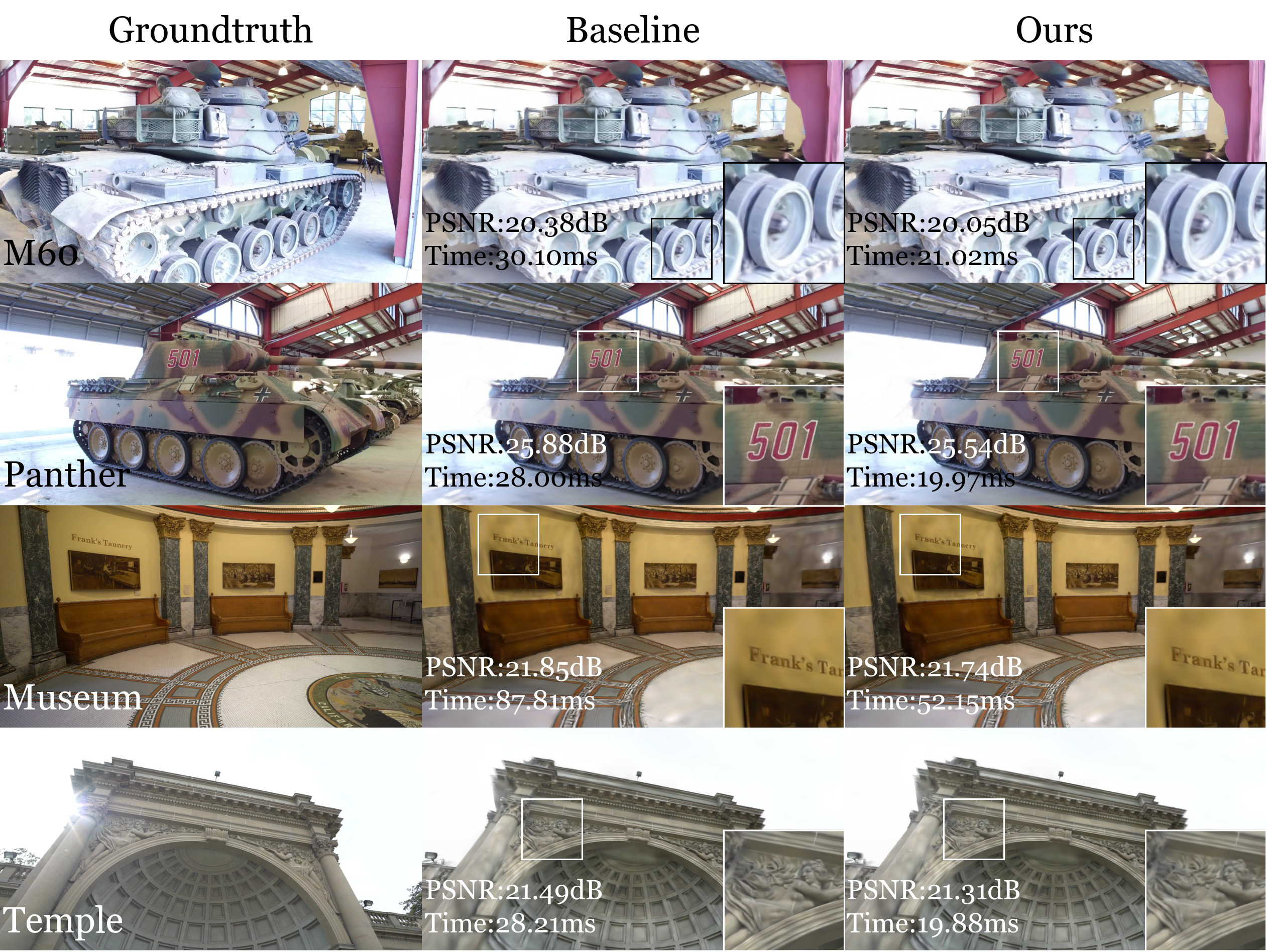}
    \caption{Visual comparison on the Tanks and Temples dataset.}
    \label{fig:vis_tnt}
\end{figure}

\begin{figure}[!htbp]
    \centering
    \includegraphics[width=0.70\textwidth]{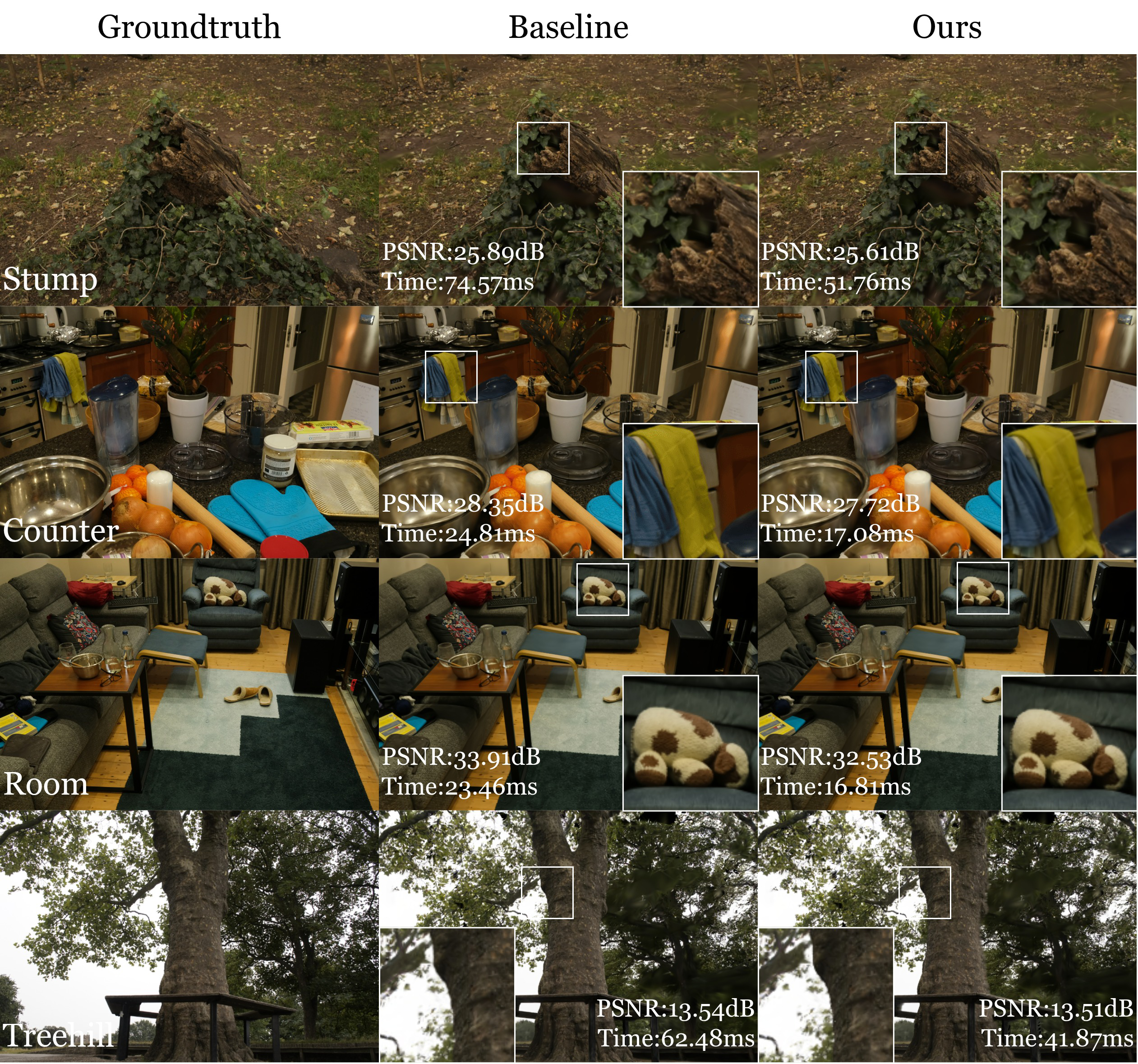}
    \caption{Visual comparison on the MipNeRF-360 dataset.}
    \label{fig:vis_mipnerf}
\end{figure}

\begin{figure}[!htbp]
    \centering
    \includegraphics[width=0.86\textwidth]{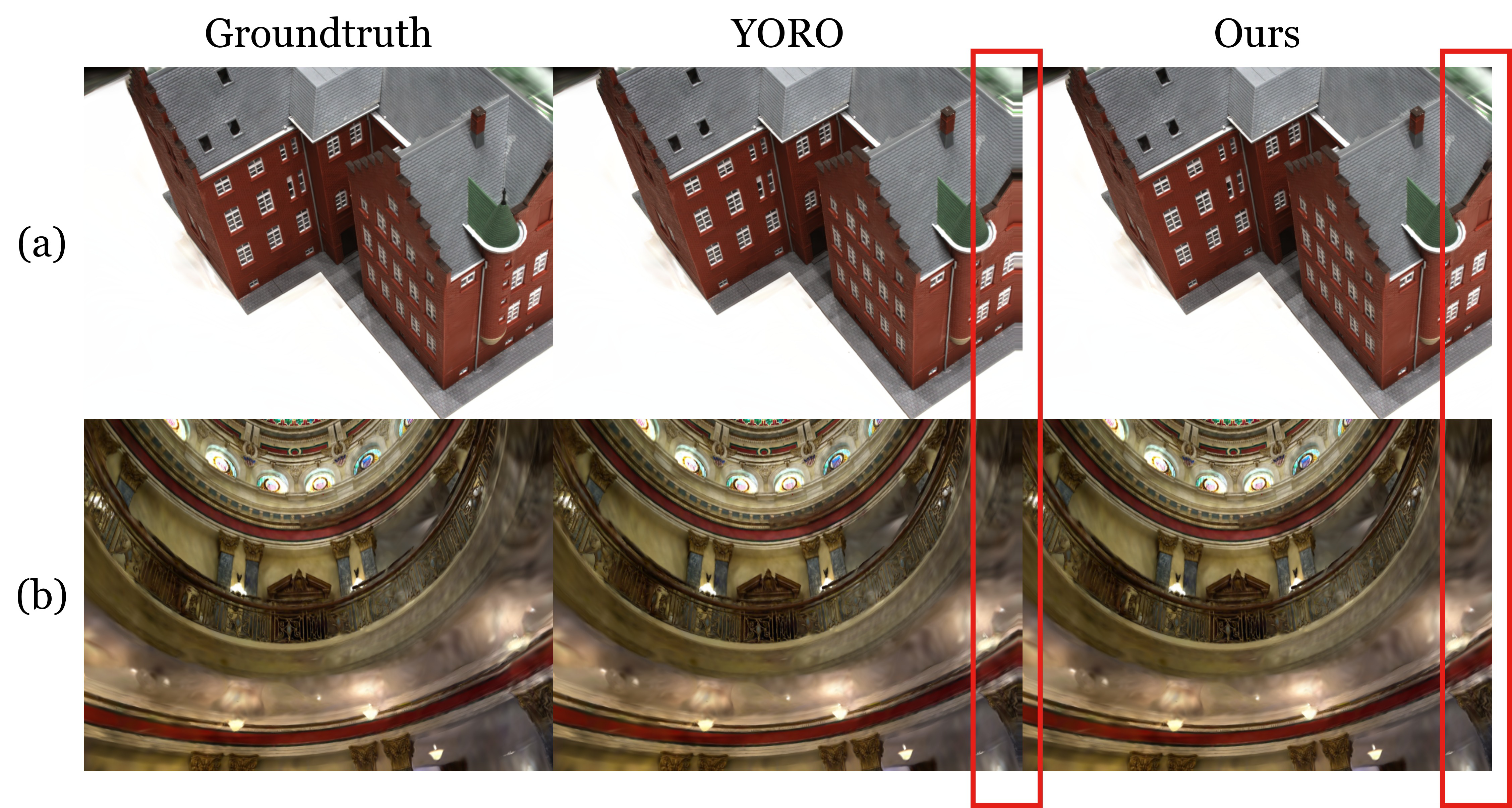}
    \caption{Qualitative comparison with YORO. (a) Near objects, where reprojection-only completion produces visible distortion. (b) Far objects, where the distortion is less visible.}
    \label{fig:yoro}
\end{figure}

\end{document}